\documentclass[conference]{IEEEtran}
\IEEEoverridecommandlockouts
\usepackage{circuitikz,adjustbox}

\usepackage{cite}
\usepackage{amsmath,amssymb,amsfonts}
\usepackage{algorithmic}
\usepackage{graphicx}
\usepackage{textcomp}
\usepackage{xcolor}

\usepackage{algorithm}
\usepackage{algorithmic}
\usepackage{svg}
\usepackage[a4paper, total={184mm,239mm}]{geometry}

\def\BibTeX{{\rm B\kern-.05em{\sc i\kern-.025em b}\kern-.08em
\kern-.1667em\lower.7ex\hbox{E}\kern-.125emX}}

\begin{document}

\title{LABIIUM: AI-Enhanced Zero-configuration Measurement Automation System \\

\thanks{{Emmanuel Olowe is sponsored by Keysight Technologies, United States. Royal Society partly supported computing resources grant RGS\textbackslash R2\textbackslash 222007.}}
}

\author{\IEEEauthorblockN{{Emmanuel A. Olowe}}
\IEEEauthorblockA{\textit{{School of Engineering}} \\
\textit{{The University of Edinburgh}}\\
{Edinburgh, UK} \\
{e.a.olowe@sms.ed.ac.uk}}
\and
\IEEEauthorblockN{{Danial Chitnis}}
\IEEEauthorblockA{\textit{{School of Engineering}} \\
\textit{{The University of Edinburgh}}\\
{Edinburgh, UK} \\
{d.chitnis@ed.ac.uk}}
}

\maketitle

\begin{abstract}
The complexity of laboratory environments requires solutions that simplify instrument interaction and enhance measurement automation. Traditional tools often require configuration, software, and programming skills, creating barriers to productivity. Previous approaches, including dedicated software suites and custom scripts, frequently fall short in providing user-friendly solutions that align with programming practices.

We present LABIIUM, an AI-enhanced, zero-configuration measurement automation system designed to streamline experimental workflows and improve user productivity. LABIIUM integrates an AI assistant powered by Large Language Models (LLMs) to generate code. LABIIUM’s Lab-Automation-Measurement Bridges (LAMBs) enable seamless instrument connectivity using standard tools such as VSCode and Python, eliminating setup overhead.

To demonstrate its capabilities, we conducted experiments involving the measurement of the parametric transfer curve of a simple two-transistor inverting amplifier with a current source load. The AI assistant was evaluated using different prompt scenarios and compared with multiple models, including Claude Sonnet 3.5, Gemini Pro 1.5, and GPT-4o. An expert solution implementing the Gradient-Weighted Adaptive Stochastic Sampling (GWASS) method was used as a baseline.

The solutions generated by the AI assistant were compared with the expert solution and a uniform linear sweep baseline with 10,000 points. The graph results show that the LLMs were able to successfully complete the most basic uniform sweep, but LLMs were unable to develop adaptive sweeping algorithms to compete with GWASS. The evaluation underscores LABIIUM’s ability to enhance laboratory productivity and support digital transformation in research and industry, and emphasizes the future work required to improve LLM performance in Electronic Measurement Science Tasks.
\end{abstract}


\section{Introduction}
Rapid advancement in technology has led to increasingly complex laboratory environments, where diverse instrumentation and sophisticated experimental setups have become the norm\cite{Skamo2023AdvantagesAcademia}. Researchers and engineers are often challenged by the need to interact with a multitude of instruments, each requiring specific configurations and proprietary software interfaces \cite{Zhao2024AskCommunity}. Traditional measurement automation tools, such as LabVIEW~\cite{Kodosky2020LabVIEW} and MATLAB\cite{TheMathWorks2024MATLAB}, while powerful, often demand extensive configuration and specialized programming skills \cite{Csokmai2021ComparativeLanguages}. These requirements can create significant barriers to productivity, hinder the pace of experimentation, and limit accessibility for those not well-versed in these platforms.

Moreover, the integration of laboratory instruments with modern programming environments remains a persistent challenge. The lack of seamless connectivity and the necessity for manual setup can lead to inefficiencies and steep learning curves, particularly for those more familiar with contemporary development tools like Python and integrated development environments (IDEs) such as Visual Studio Code \cite{Kumar2023PerformanceLanguages}. Previous attempts to address these issues through dedicated software suites or custom scripting have often fallen short, failing to provide flexible, user-friendly solutions that align with current programming practices \cite{Rao2024DevelopmentProtocol, Barnes2023SwiftVISA:Protocol}.

There is a pressing need for innovative solutions that simplify instrument interaction, enhance measurement automation, and integrate effortlessly with modern software tools. The incorporation of artificial intelligence (AI), specifically large language models (LLMs)\cite{Brown2020LanguageLearners}, presents a promising avenue to revolutionize laboratory automation by enabling intelligent code generation and error correction\cite{Wang2024ExecutableAgents}. There is already work focusing on using LLMs to generate code for SCPI Compatible measurement instruments\cite{Febba2024FromInstruments}. This can be further enhanced through the pathway of tool use\cite{Schick2023Toolformer:Tools, Patil2023Gorilla:APIs}, this is where LLMs are able to invoke the use of external application programming interfaces (APIs) \cite{Shen2024LLMSurvey} allowing the LLMs to interact directly with the instruments.

To address these challenges, we introduce LABIIUM, an AI-enhanced, zero-configuration measurement automation system designed to streamline experimental workflows and significantly boost user productivity. LABIIUM leverages an AI assistant built upon LLMs to generate code for measurement tasks, assist in error correction, and accelerate prototyping. This integration of AI not only reduces the technical overhead associated with instrument control but also democratizes access for users with varying levels of programming expertise.

Central to LABIIUM are the Lab-Automation-Measurement Bridges (LAMBs), devices that facilitate seamless connectivity to laboratory instruments using standard tools like Visual Studio Code and Python, without the need for any setup or configuration. By implementing a custom Virtual Instrument Software Architecture (VISA) \cite{Foundation2024VirtualSpecification} and driver system, LABIIUM allows users to interact with instruments through familiar programming paradigms, thus bridging the gap between complex hardware interfaces and user-friendly software environments. In addition, LAMBs provide a standardized environment, in which LLMs can then use through tool-use to invoke functionality and give code suggestions given the user's instrument configurations for their measurement.  

To verify the efficacy of LABIIUM, we conducted experiments involving the measurement of the parametric transfer curve of a simple two-transistor inverting amplifier with a current source load.

The AI assistant was evaluated using different prompt scenarios and compared with multiple underlying models, including Claude Sonnet 3.5, Gemini Pro 1.5, and GPT-4o. An expert solution implementing the Gradient-Weighted Adaptive Stochastic Sampling (GWASS) method was used as a benchmark.




\section{Methodology}

\label{sec:methodology}
In this section, we describe the experimental setup, the AI models used, the prompt scenarios, and the methods employed to evaluate the efficacy of LABIIUM in automating measurement tasks.

\subsection{Lab Automation Measurement Bridge (LAMBs)}
\begin{figure*}[t]
    \centering
    \includegraphics[width=\textwidth]{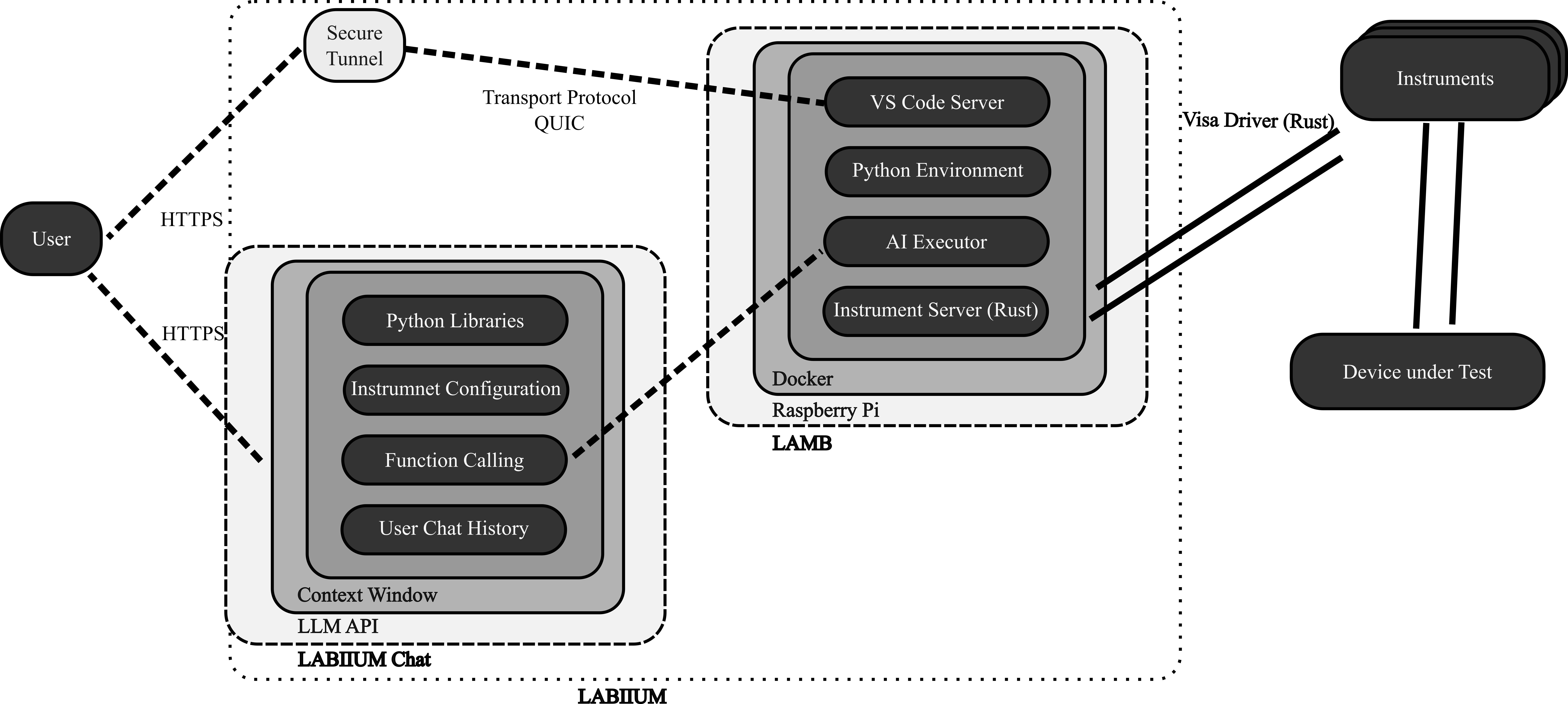}
    \caption{Architecture of the Lab-Automation Measurement Bridge (LAMB). The Raspberry Pi4 serves as the standardized hardware platform, running the Rust-based VISA implementation and server. The system interfaces with laboratory instruments via the USBTMC protocol and supports connections through model names or serial numbers. Users interact with the LAMB through a browser-based VS Code and Python environment, utilizing an abstraction layer that simplifies SCPI command usage. Connectivity is available through a secure tunnel connections over the internet.}
    \label{fig:lamb_architecture}
\end{figure*}

The Lab-Automation Measurement Bridges (LAMBs) are the keystone of the LABIIUM system, providing a seamless and standardized interface between laboratory instruments and the automation framework. Designed to facilitate efficient and flexible instrument management, LAMBs integrate several key components and technologies, each contributing to the overall functionality and ease of use. The architecture of a LAMB is illustrated in Figure~\ref{fig:lamb_architecture}.

Each LAMB is built upon a Raspberry Pi4, chosen for its balance of computational power, connectivity options, and affordability. The Raspberry Pi4 provides a reliable and consistent hardware foundation, ensuring that each bridge can handle multiple instruments and sustain continuous operation within various laboratory environments. 

At the heart of LAMB's instrument communication lies a VISA (Virtual Instrument Software Architecture) implementation developed in the Rust Programming Language. This choice leverages Rust's performance and safety features, ensuring robust and efficient interactions with connected instruments. LAMB utilizes the USB Test and Measurement Class (USBTMC) protocol to communicate with instruments over USB. USBTMC is a standardized protocol that facilitates high-speed data transfer and reliable command execution, making it ideal for laboratory measurements. The Rust VISA implementation allows connections to instruments either by their model names or by their unique serial numbers. This flexibility ensures that users can easily identify and manage instruments, even in environments with multiple devices of the same type.

To enable remote and concurrent command execution, each LAMB runs a Rust-based server that interfaces with the VISA layer. This server provides several critical functionalities. Users can send commands to instruments remotely, eliminating the need for physical access and allowing for distributed laboratory setups enabling remote command execution. The server accepts Standard Commands for Programmable Instruments (SCPI)~\cite{SCPIConsortium1999StandardStyle}, a widely adopted command language in test and measurement equipment. By supporting SCPI, LAMB ensures compatibility with a vast array of instruments. The server can handle multiple instruments simultaneously, managing their individual command queues and ensuring synchronized operations across devices.


LAMBs are designed to integrate seamlessly with modern development environments, specifically Visual Studio Code (VS Code) and Python. This integration is achieved through an abstraction layer that simplifies interaction with instrument. Users can write and execute Python scripts within VS Code to control instruments connected via LAMB. This setup leverages the extensive ecosystem of Python libraries and VS Code's powerful development tools. 

LAMBs offer versatile connectivity options to accommodate various laboratory setups. Users can connect to a LAMB directly via an Ethernet cable, providing a stable and high-speed local network connection. For remote access, LAMB supports secure tunnel connections over the internet. This feature enables users to interact with instruments from anywhere, facilitating collaborative research and distributed laboratory operations.

The operational workflow of a LAMB involves several stages. Upon startup, the Raspberry Pi4 initializes the Rust VISA implementation and establishes connections with all recognized instruments based on model names or serial numbers. The Rust server listens for incoming SCPI commands from remote clients or local scripts executed within the VS Code environment. Received commands are parsed and forwarded to the appropriate instrument via the USBTMC interface. Responses from instruments are relayed back to the requesting entity. Measurement data is collected, processed, and stored as needed. Users can retrieve and visualize this data through their development environment.

By encapsulating these functionalities within a standardized and modular framework, LAMBs provide a robust foundation for automated measurement tasks. This design not only simplifies instrument management but also enhances the scalability and adaptability of laboratory automation systems.

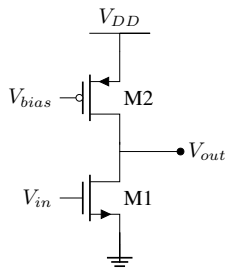
\begin{figure}[t]
    \centering
    
    \begin{adjustbox}{scale=0.8}
    \begin{circuitikz}[american voltages]{
    
    \ctikzset{tripoles/mos style/arrows}
    \ctikzset{tripoles/pmos style/emptycircle}
    \draw 
    (1,0) to[short] ++(0,0.5) node[ground] {}
    (1,0) to
    (1,0.3) node[nmos,anchor=S,emptycircle](M1){M1} 
    (M1.drain) to (1,2)
    (M1.drain) -- + (1,0) node[circ] {} node[right] {$V_{out}$}
    (M1.gate) node[left] {$V_{in}$}
    (1,3.5) node[pmos,anchor=S,emptycircle](M2){M2}
    (M2.source) -- + (0,0.3)  coordinate (vdd)
    (vdd) -- +(0.5,0) -- ++(-0.5,0)
    (vdd) node[above] {$V_{DD}$}
    (M2.gate) node[left] {$V_{bias}$};
    }

    \end{circuitikz}
    \end{adjustbox}

   \caption{Schematic diagram of the two-transistor inverting amplifier with a current source load. The input signal \( V_{\text{in}} \) is applied to the gate of \( M1 \) (BS170), an N-channel, which serves as the amplifying transistor. The drain of \( M1 \) is connected to the source of the current source load \( M2 \) (BS250P), a P-channel, whose gate is biased by a fixed voltage \( V_{\text{bias}} \) and whose current can be controlled by means of changing \(V_{\text{bias}}\). The amplified output \( V_{\text{out}} \) is taken from the drain of \( M1 \) and \( M2\), and the circuit is powered by a positive supply voltage \( V_{\text{DD}} \).}
   \label{fig:circuit}
\end{figure}

\subsection{LABIIUM Chat}
\label{sec:labium_chat}

LABIIUM Chat constitutes the key interactive layer of the LABIIUM, integrating LLM capabilities directly into the user's measurement workflow. It provides a natural language interface through which users can query, configure and control their experiments bridging the gap between human instructions and the code execution required for laboratory automation. Leveraging the standardized environment created by LAMB's hardware interface LABIIUM Chat simplifies the complex instrument operations into accessible conversational interactions.

A key challenge when working with LLM APIs can be working within the confines of their effective context windows, as LLMs can forget  information deep within long contexts \cite{Liu2024LostContexts} and LLM performance may decline as large contexts are used\cite{An2024WhyShort}. In the light of needing to use large Python Library in the context to handle the programming needs of the user. LABIIUM Chat address this by employing a dedicated rust library to convert python libraries into appropriate context for LLMs. This involves optimizations such as the omission of the code within methods and functions and only the use of the doc strings, selective omission of unnecessary components of the library. YAML configurations can be made to contain and define these necessary optimizations or minimum contexts to reduce the various libraries token consumption. 

Beyond static code generation, LABIIUM Chat incorporates function calling functionality to enhance the LLM’s role as an intelligent measurement controller. Recent advancements in LLM tool use\cite{Schick2023Toolformer:Tools, Patil2023Gorilla:APIs} allow the model to invoke predefined Python functions that directly interface with LAMBs. By exposing a set of carefully designed commands—ranging from sending SCPI instructions to the connected instruments to collecting measurement data—LABIIUM Chat grants the LLM the agency to execute actions autonomously. For example, if a user requests a parametric sweep or an adaptive sampling routine such as those presented in this paper, the LLM can internally call Python functions to adjust output voltages, trigger measurements, and retrieve results, all without leaving the chat interface. This also gives LLMs the ability to recover and self-correct encounter programming errors as they receive all the output of the code execution.

\begin{figure*}[t]
    \centering
    \includegraphics[width=1\textwidth]{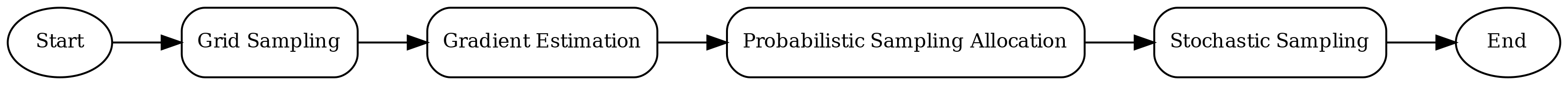}
    \caption{Gradient-Weighted Adaptive Stochastic Sampling Algorithm Process}
    \label{fig:gwass}
\end{figure*}

\begin{table*}[t]
\centering
\renewcommand{\arraystretch}{1.3}
\begin{tabular}{|c|p{4cm}|p{4cm}|p{7cm}|}
\hline
\textbf{Scenario} & \textbf{Description} & \textbf{Expected Outcome} & \textbf{Observed Outcome} \\
\hline
\textbf{Experiment 1} & The AI system is asked to generate and execute code to perform an efficient sweep and generate a graph of the transfer curve. & The AI should generate a measurement script that efficiently sweeps the input voltage and records output voltage to produce a transfer curve. & The AIs produced a uniform sweep, failing to optimize the sampling for regions with high variability, but results match largely the reference uniform sweep. \\
\hline
\textbf{Experiment 2} & The AI system is asked to sample from regions with the highest rates of change, aiming to implement an adaptive sampling method. & The AI should concentrate samples around the transition region of the inverter to improve resolution. & The AIs did not properly implement adaptive sampling and continued to use a uniform approach, with no notable variation from the base uniform sweep reference. \\
\hline
\textbf{Experiment 3} & The AI system is given a full description of the GWASS method and asked to use it to sample and generate the measurement code. & The AI should correctly implement GWASS, placing more samples in high-gradient regions to improve efficiency. & Overall these results massively decline in performance relative to other outcomes. They fail to fully implement GWASS and remained largely uniform and fail to deliver the correct number of points, with one model failing entirely. \\
\hline
\end{tabular}
\vspace{0.5em} 
\caption{Comparison of LLM Performance Across Three Experiment Scenarios}
\label{tab:llm_experiments}
\end{table*}

\subsection{Efficient Sampling Problem}
Efficient sampling in graph generation is critical for capturing key features of measurements while minimizing time and resource use. Adaptive methods like Gradient-Weighted Adaptive Stochastic Sampling (GWASS) focus sampling in regions with high variability, reducing redundant data collection and enhancing precision with fewer points. By incorporating such strategies, LABIIUM showcases the potential of large language models (LLMs) to automate complex, intelligent measurement tasks. Using GWASS as a benchmark, the experiments highlight both the promise and current limitations of LLMs, emphasizing their potential to enhance laboratory efficiency and optimize experimental workflows. An expert solution implementing the experiment with the GWASS method was developed to serve as a benchmark. Additionally, a uniform sweep solution with 10,000 points was performed to serve as a baseline for comparison. This extraction of high resolution data requires more than 12 hours with the use of a Keysight EDU36311A Power Supply Unit (PSU) and a Keysight EDU34450A 5 $(\frac{1}{2})$ Digit Multimeter (DMM) used in the experiment to measure the circuit. The measurements obtained using the DMM exhibited a standard deviation of \(8.58 \times 10^{-7}\), reflecting the instrument's high precision and minimal variability in voltage readings. The measurement setup involves controlling the PSUs to supply specific voltages and measuring the output voltage \( V_{\text{out}} \) using the multimeter. Channel 1 of the PSU was swept, \( V_{\text{in}} \) from 0\,V to 5\,V with a maximum current of 0.1\,A. Channel 2 of the PSU was set \( V_{\text{DD}} \) to 3\,V with a maximum current of 0.1\,A. Channel 3 of the PSU was swept \( V_{\text{bias}} \) from 0\,V to 5\,V with a maximum current of 0.1\,A. The DMM then then measured \( V_{\text{out}} \). For each value of \( V_{\text{bias}} \) in the range of 0\,V to 5\,V, AI-generated would then sample up to 100 points of \( V_{\text{in}} \). The expert GWASS implementation and the uniform sweep baseline collected data accordingly for comparison.

\subsection{AI Models and Prompt Scenarios}
To evaluate LABIIUM's AI assistant capabilities, we compared its performance with three underlying Large Language Models (LLMs): GPT-4o (OpenAI), Claude Sonnet 3.5 (Anthropic) and Gemini Pro 1.5 (Google). 

For each LLM, we tested three prompt scenarios: Experiment 1: The AI system is asked to generate and execute code to perform an efficient sweep and generate a graph of the transfer curve. Experiment 2: The AI system is asked to sample from regions with the highest rates of change, aiming to implement an adaptive sampling method. Experiment 3: The Gradient-Weighted Adaptive Stochastic Sampling (GWASS) method is described completely, and the AI system is asked to use it to sample and generate the measurement code. The output of all scenarios is limited to sample or sweep only 100 points across \(V_{\text{in}}\) and 10 points across \( V_{\text{bias}} \).

\begin{figure*}[!ht]
    \centering
    \includegraphics[width=0.9\textwidth]{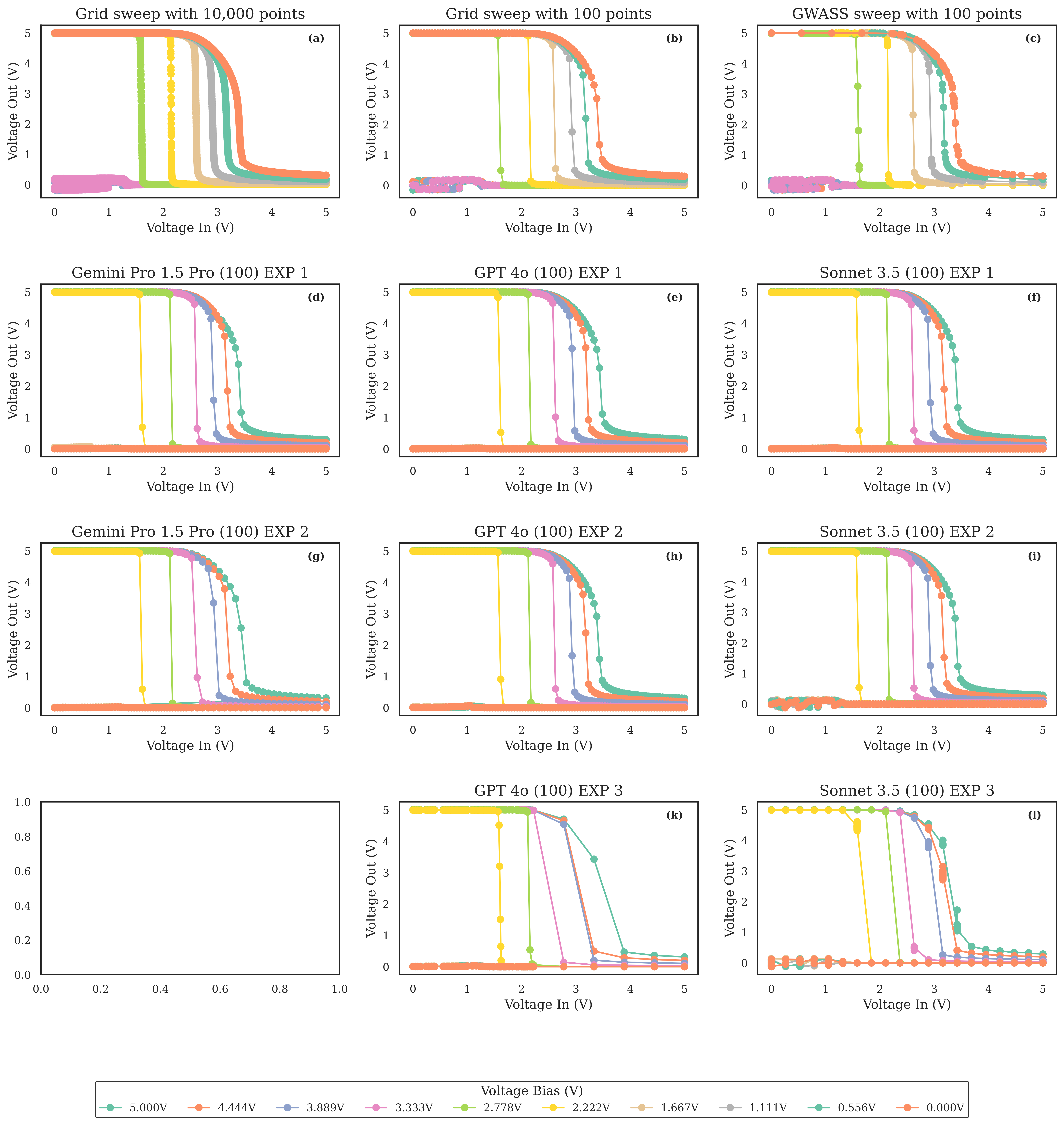}
    \caption{(a), (b) represent linear sweeps of differing resolution, with (a) employing 10,000 points and (b) using only 100 points. (c) illustrates the GWASS method, focusing samples around the inverter’s switching region for enhanced detail with the same number of points as (b). (d), (e), (f) show results from Experiment 1 attempts using LLM-generated methods, which remain essentially uniform and do not adaptively concentrate samples. (g), (h), (i) correspond to Experiment 2 attempts, again lacking adaptive refinement. (j), (k), (l) present Experiment 3 attempts where the LLMs were given a full description of GWASS; while slightly improved, these still fail to match the expert’s adaptive efficiency seen in (c). Different line colors represent varying $V_{\text{bias}}$ values, illustrating how the inverter’s transition flattens and shifts as $V_{\text{bias}}$ decreases.}
    \label{fig:transfer_curve}
\end{figure*}

\subsection{Gradient-Weighted Adaptive Stochastic Sampling (GWASS)}

To enhance sampling efficiency in regions where the circuit's response changes rapidly, we employed the GWASS method.  GWASS operates in two main phases. First an initial coarse sampling is performed, this is where a small fraction of the total allowed evaluations is used to perform a coarse sampling of the parameter space. This provides an initial estimate of the gradient magnitudes across the domain. The remaining evaluations are allocated probabilistically to regions with higher estimated gradients. Sampling within these regions is performed uniformly but with a higher density, concentrating computational resources where they are most needed. Figure~\ref{fig:gwass}

\section{Results}
\label{sec:results}

The inverter under consideration exhibits a characteristic transfer: at low input voltages, the output remains high; as the input voltage (\(V_{\text{in}}\)) surpasses a certain threshold, the output (\(V_{\text{out}}\)) transitions to a lower level. Reducing the bias voltage (\(V_{\text{bias}}\)) shifts this threshold and reduces the slope of the transition, thereby lowering the effective gain near the inverter’s switching point.

Figure~\ref{fig:transfer_curve} presents a series of transfer curves for various \(V_{\text{bias}}\) values under different measurement strategies and conditions. Subplots (a) and (b) illustrate uniform (linear) sweeps. Subplot (a) uses 10,000 points, providing a high-resolution reference, while (b) employs only 100 points, capturing the overall shape but with reduced detail. In both cases, the expected inverter behavior is evident: a steep transition at higher \(V_{\text{bias}}\) values that flattens and shifts as \(V_{\text{bias}}\) decreases.

In (c), the Gradient-Weighted Adaptive Stochastic Sampling (GWASS) method is shown. With only 100 points, GWASS concentrates sampling effort near the steep transition, effectively highlighting the critical switching region. This stands in contrast to the uniform approaches, where data points are spread evenly and do not emphasize regions of rapid change.

Subplots (d), (e), and (f) correspond to Experiment 1 (EXP 1) results from different Large Language Model (LLM)-guided attempts. Although these LLM-based approaches produce executable measurement code, they fail to emulate adaptive sampling. The resulting curves resemble the simple uniform strategy of (b), without improved localization of the threshold region.

In (g), (h), and (i), corresponding to Experiment 2 (EXP 2), the LLMs were given more specific instructions. Nonetheless, their outputs remain uniform-like, showing no marked improvement over EXP 1 attempts. Finally, in (j), (k), and (l), representing Experiment 3 (EXP 3), the LLMs received a full description of the GWASS method. While these attempts occasionally place a few points more strategically, they still do not achieve the level of adaptive efficiency seen in (c). In other words, none of the LLM-based strategies match the expert’s ability to focus on the steepest part of the transition curve using limited samples.

Overall, these results confirm that while uniform sweeps and LLM attempts capture the general inverter characteristic, only the expert GWASS approach successfully adapts to the dynamic gradient of the transfer curve, efficiently highlighting the transition region and outperforming naive sampling distributions.

\section{Discussion}
\label{sec:discussion}

The presented findings underscore both the potential and limitations of integrating LLMs into laboratory automation workflows. On one hand, LLM-guided solutions can readily produce functional code that executes measurement tasks without extensive manual setup. This capability lowers the barriers to entry for complex experiments and can expedite routine data collection.

On the other hand, these models struggle to implement more sophisticated, context-sensitive strategies. While GWASS efficiently allocates samples to the steepest portions of the transition curve, the LLM-generated attempts remain essentially uniform, demonstrating limited capacity for dynamic decision-making. Achieving true adaptive sampling requires more than just describing an algorithm in text form; it necessitates the model’s deeper understanding of underlying principles and its ability to iteratively adjust measurements based on real-time feedback.

Improving LLM performance may involve specialized training on domain-relevant datasets, enhanced prompt engineering, and integrating external state-management tools that allow the model to retain and act upon measurement history. Such refinements could enable future AI-driven measurement systems to not only automate routine experiments but also optimize their sampling strategies, pinpointing critical operating regions with fewer data points.

In short, while current LLM solutions serve as a useful starting point, bridging the gap between uniform sweeps and truly adaptive sampling approaches like GWASS remains an open challenge. As LLMs and their associated tool-use capabilities improve, we can anticipate more intelligent, resource-efficient measurement methodologies emerging in both research and industrial settings.

\section{Conclusion}
\label{sec:conclusion}
In this paper, we introduced LABIIUM, an AI-enhanced, zero-configuration measurement automation system designed to streamline experimental workflows and improve productivity. By integrating LLM-powered AI assistants within the standardized environment provided by LAMBs, LABIIUM enables efficient measurement automation and facilitates advanced sampling techniques.

Our experiments demonstrated that LABIIUM, leveraging AI and the GWASS method, achieved measurement efficiencies comparable to the human-expert baseline while reducing setup complexity. The standardized and modular framework of LABIIUM played a pivotal role in bridging the gap between sophisticated instrument interfaces and user-friendly programming environments, making AI-assisted measurement tasks accessible and highly productive.

These findings underscore LABIIUM’s potential to enhance laboratory workflows and align with the growing need for automation in research and industry. Future work will focus on extending its capabilities with more advanced AI models and sampling methodologies to further drive productivity and innovation in experimental environments.

\section*{Acknowledgements}

{The authors like to thank Keysight Technologies, United States, for their funding support of this project, Royal Society grant RGS\textbackslash R2\textbackslash 222007 for their support in computing resources, EDINA and ISG@University of Edinburgh for their support in enabling access to OpenAI services, and the technical staff of Kivlin Suit: Iain Gold, Alan Robertson, and Alasdair Christie for their lab support.}

\bibliographystyle{IEEEtran}
\bibliography{IEEEabrv,references}
\vspace{12pt}
\end{document}